\documentclass{article}

\usepackage{amsmath,graphicx,mlspconf}

\usepackage{xcolor}
\usepackage{amsfonts}
\usepackage{amssymb}
\usepackage{multirow}
\usepackage{booktabs}

\usepackage[hidelinks]{hyperref}

\title{Learning with Volterra Neural Networks: A System Theoretic Perspective}

\name{
Haoyu Yun\textsuperscript{1},
Hamid Krim\textsuperscript{1},
Yufang Bao\textsuperscript{2}%
\thanks{
This work was supported by the U.S. Army Research Office (ARO)
under Grant No.~W911NF2410329.
}
}

\address{%
\textsuperscript{1}Department of Electrical and Computer Engineering,
North Carolina State University, Raleigh, NC, USA\\
\textsuperscript{2}Department of Mathematics and Computer Science,
Fayetteville State University, Fayetteville, NC, USA
}

\begin{document}

\maketitle

\begin{abstract}
Higher-order interaction components are important for signal, image, and video modeling, but explicit high-order operators often suffer from rapidly increasing parameter and computational costs. This paper presents kVNN, a learnable kernelized Volterra Neural operator for compact higher-order filtering. The motivation is to use kernelization to improve the efficiency of Volterra-type neural operators while providing a structured interpretation of their higher-order components. The proposed formulation combines the order-wise structure of Volterra filtering with learnable polynomial-kernel atoms, allowing different interaction orders to be represented by separate learnable centers and coefficients. This order-decoupled representation avoids explicit high-order tensor parameterization and can be implemented as a CNN-compatible layer. Experiments on representative vision tasks show that kVNN achieves a favorable accuracy--efficiency trade-off.
\end{abstract}


\newcommand{\cem}[1]{\textcolor{blue}{cem: #1}}

\section{Introduction}
\label{sec:introduction}

Deep neural networks have achieved remarkable performance in signal, image, and video analysis. Their core building blocks usually consist of linear filtering operators and pointwise nonlinearities. Although this design is effective, the local filtering operators themselves are mostly linear, and higher-order interactions among input variables, channels, and local structures are often modeled and implicitly captured by cross-layer entanglement through network depth. For signal processing tasks with complex local dependencies, such as video motion patterns, image textures, and multi-scale feature couplings, explicit higher-order modeling can improve representation capacity. This observation constitutes the main motivation for this work: to develop an efficient higher-order filtering operator that improves representation capacity while keeping model complexity controlled.

Volterra filtering provides a classical framework for nonlinear system modeling through order-specific interaction components. Recent Volterra-inspired neural models have shown that explicit higher-order representations can improve expressive power, but directly parameterizing high-order Volterra tensors causes the number of parameters and computations to grow rapidly with interaction order and input dimension~\cite{JMLR:v25:21-1082}. Kernel methods provide a principled way to represent polynomial-type higher-order interactions without explicitly constructing high-dimensional monomial features~\cite{Franz:2006}. The motivation of this work is to, therefore, use kernelization to improve the efficiency of Volterra-type neural operators and to provide further insight into their functional structure. However, classical kernel expansions usually rely on training-sample-based representations, whose complexity grows with the number of samples, making them unsuitable as reusable neural operators in deep networks.

Existing kernelized convolutional methods have attempted to introduce kernels into convolutional operations. However, a direct use of kernel tricks in local filtering does not necessarily provide a structured Volterra-type neural operator. For example, Kervolution~\cite{Wang2019CVPR} introduces kernel functions into convolutional layers, but its kernel form is usually pre-defined and is not explicitly organized according to the order-wise decomposition of Volterra filtering. As a result, interactions of different orders can be coupled within the same kernelized representation, which limits the independent parameterization and optimization of different interaction orders.

To address these issues, this paper proposes a novel learnable kernelized Volterra Neural operator, referred to as kVNN. The key idea is to combine the order-wise decomposition structure of Volterra filtering with learnable polynomial-kernel atoms. Unlike directly learning full high-order Volterra tensors or using a fixed-form kernelized convolution, kVNN assigns independent learnable kernel centers and coefficients to different interaction orders, forming a structured multi-kernel Volterra representation. This design explicitly decouples linear, quadratic, and higher-order interactions while maintaining a compact parameterization. The proposed representation is further implemented as a convolution-like layer, enabling it to replace conventional convolutional filters while preserving CNN-style channel organization.

The main contributions of this paper are summarized as follows. First, a learnable multi-kernel Volterra representation is introduced to compactly model higher-order interactions using order-specific polynomial-kernel atoms. Second, formal approximation results are provided to interpret kVNN cells, layers, and multilayer networks from a local-to-global geometric perspective. Finally, the proposed operator is implemented as a CNN-compatible layer and evaluated on video action recognition, image denoising, and image classification tasks. Experimental results show that kVNN achieves competitive or improved performance while maintaining low model complexity, demonstrating a favorable accuracy--efficiency trade-off.

\section{Background}
\label{sec:background}

\subsection{Volterra Filtering and Higher-Order Modeling}

Volterra series provides a classical framework for nonlinear system modeling and can be viewed as a higher-order extension of linear filtering. For a finite-dimensional input $x\in\mathbb{R}^{d}$, a truncated Volterra model represents the response as a sum of order-specific components. The first-order term corresponds to a linear response, while second- and higher-order terms describe nonlinear couplings among input variables. This makes Volterra filtering a natural mathematical basis for constructing higher-order local operators in neural networks. However, the $r$-th order Volterra component is generally represented by an $r$-way coefficient tensor, whose number of parameters grows rapidly with both the input dimension and the interaction order. This tensor parameterization makes explicit high-order Volterra filters difficult to use directly as scalable and reusable layers in modern deep architectures.

Higher-order interaction modeling has also been explored in visual and signal representation learning. Bilinear models use second-order feature interactions to improve fine-grained visual recognition~\cite{7410527}. Polynomial neural networks and tensor-based mappings further enhance nonlinear representation capacity from the perspectives of polynomial and tensor representations~\cite{9156685,10.5555/3455716.3455839}. Volterra-inspired higher-order convolutional architectures have also been studied for image classification and structured representation learning~\cite{9247263,JMLR:v25:21-1082}. These studies show that explicit higher-order representations can improve model expressivity. However, many existing higher-order designs are introduced as specialized representation modules or task-specific components, rather than compact filtering operators that can directly replace standard convolutional layers across different network backbones.

\subsection{Kernelization and Kernelized Convolution}

Kernel methods provide a principled way to represent polynomial type interactions without explicitly constructing all high-dimensional monomial features. In particular, polynomial kernels are naturally connected to Volterra components because they implicitly encode homogeneous polynomial interactions of different degrees~\cite{Franz:2006}. This connection suggests that kernelization can reduce the complexity of high-order Volterra modeling while preserving a clear functional interpretation. However, classical kernel methods usually rely on sample-centered expansions. Although such representations are theoretically well understood, their size grows with the number of training samples. This makes them unsuitable as reusable neural operators in deep networks, where a layer should have a fixed set of trainable parameters independent of the training set size and should be optimized end-to-end with the task loss.

Kernelized convolutional methods show that kernel functions can be introduced into local convolutional filtering. For example, Kervolution~\cite{Wang2019CVPR} replaces standard convolution with a kernelized operation. However, such methods mainly apply a pre-defined kernel trick to convolutional filtering, rather than deriving the operator from the order-wise structure of Volterra filtering. Their kernel forms are usually fixed in advance, and different interaction orders may be coupled within the same representation. This limits the independent parameterization and optimization of different orders.

These limitations motivate a learnable Volterra-kernel formulation in which different interaction orders are represented by compact polynomial-kernel atoms with learnable centers and coefficients. Such a formulation preserves the structural interpretation of Volterra filtering, reduces the cost of explicit tensor parameterization, and provides the basis for constructing a CNN-compatible higher-order neural filtering layer.

\section{Methodology}
\label{sec:Method}
In this section, we first introduce the design of the kVNN operator, which represents order-wise Volterra interactions through learnable polynomial-kernel atoms without explicitly learning high-order tensors. We then interpret this operator at different architectural levels, from a single kVNN cell to a kVNN layer and finally to a multilayer kVNN network.

\begin{figure}[t]
    \centering
    \includegraphics[width=0.95\linewidth]{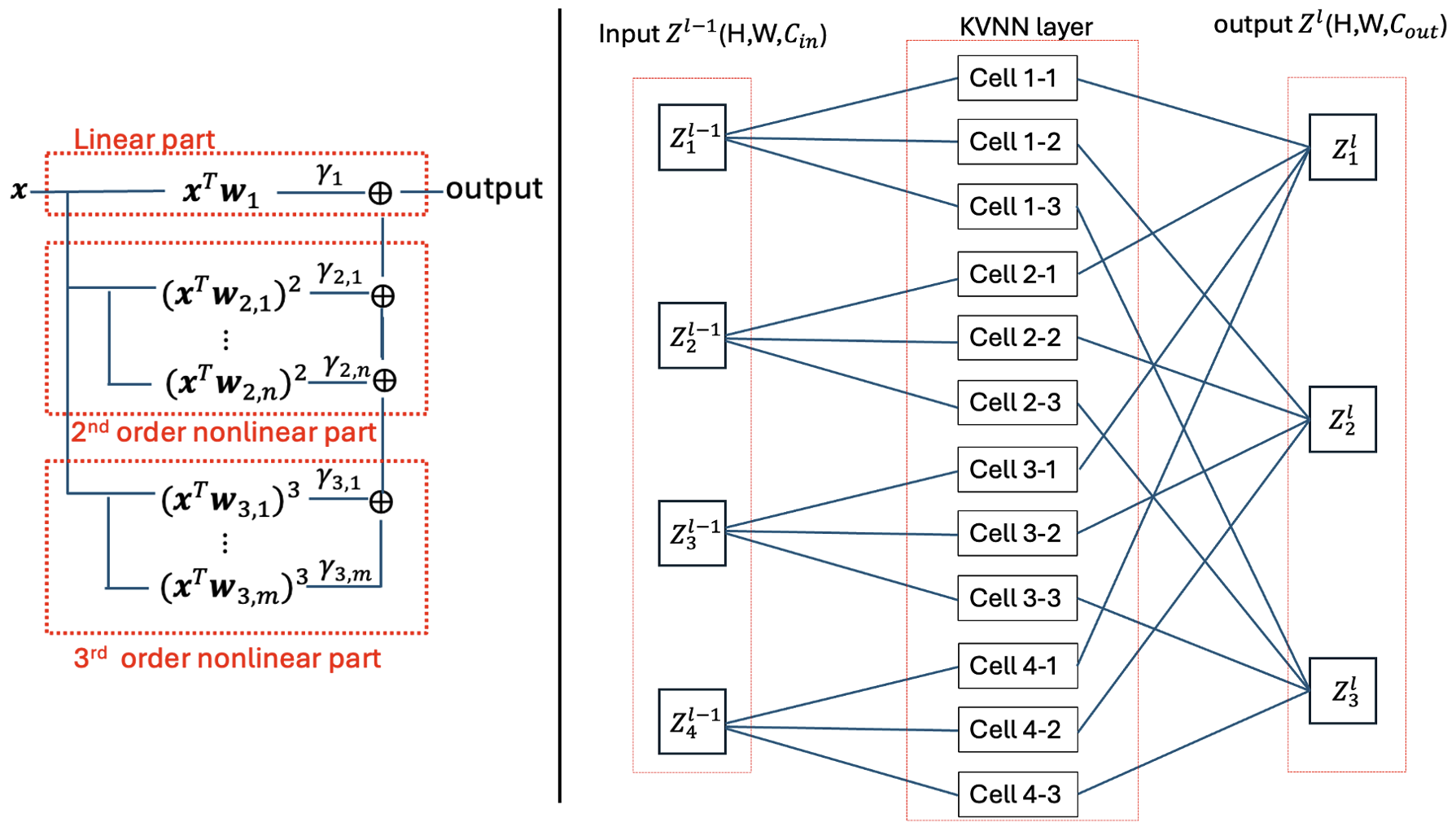}
    \caption{
    Proposed kVNN cell and layer structure. 
    A kVNN cell decomposes one nonlinear filter into order-specific learnable kernel branches, where different orders use separate centers and coefficients. 
    Multiple kVNN cells are arranged in parallel to form a kVNN layer, following a CNN-like channel organization.
    }
    \label{fig:kvnn_structure}
\end{figure}

\subsection{Learnable Multi-Kernel Volterra Representation}
\label{subsec:learnable_multikernel}

This subsection develops the mathematical formulation of the proposed kVNN operator. The key idea is to represent each Volterra order by a finite set of learnable polynomial-kernel atoms, so that different interaction orders are parameterized separately while avoiding explicit high-order tensor parameterization.

Let $x\in\mathbb{R}^{d}$ denote a finite-dimensional input signal, such as a vectorized local image patch. 
The kVNN operator is defined as an order-wise sum of learnable kernel responses:
\begin{equation}
\label{eq:kvnn_multikernel_method}
f_{\mathrm{kVNN}}(x)
=
\sum_{r=1}^{p}
\sum_{i=1}^{M_r}
\gamma_{r,i}
\left(x^\top w_{r,i}\right)^r .
\end{equation}
Here, $r$ denotes the interaction order, $M_r$ is the number of atoms used for the $r$-th order component, $w_{r,i}\in\mathbb{R}^{d}$ is a learnable kernel center, and $\gamma_{r,i}$ is the corresponding learnable coefficient. 
For each order, the component response can be written as
\begin{equation}
\label{eq:kvnn_order_component_method}
f_r(x)
=
\sum_{i=1}^{M_r}
\gamma_{r,i}
\left(x^\top w_{r,i}\right)^r ,
\end{equation}
with an order-specific set of learnable centers
\begin{equation}
\label{eq:order_specific_centers_method}
\mathcal{W}_r
=
\{w_{r,i}\}_{i=1}^{M_r}
\subset \mathbb{R}^{d}.
\end{equation}
Thus, the linear, quadratic, and higher-order responses are represented by separate learnable components rather than being coupled inside a single fixed kernel form.

This representation is designed to efficiently approximate the truncated Volterra filtering structure without explicitly parameterizing high-order Volterra tensors. 
A $p$-th order truncated Volterra mapping~\cite{Volterra:1930} takes the form
\begin{equation}
\label{eq:volterra_tensor_form_method}
f_{\mathrm{V}}(x)
=
\sum_{r=1}^{p}
\sum_{i_1=1}^{d}\cdots\sum_{i_r=1}^{d}
h_r(i_1,\ldots,i_r)
x_{i_1}\cdots x_{i_r},
\end{equation}
where $h_r$ is the $r$-th order coefficient tensor. 
This tensor explicitly captures $r$-way interactions among input coordinates. 
However, the number of coefficients in the $r$-th order component generally grows as $O(d^r)$, which makes explicit high-order Volterra filtering expensive when used as a reusable operator in deep networks.

Our method is based on kernelization of the Volterra filtering structure. 
Kernelization provides a compact way to express polynomial-type interactions without explicitly enumerating all high-order tensor coefficients or monomial features. 
For the $r$-th order component, kVNN uses the homogeneous polynomial-kernel atom
\begin{equation}
\label{eq:poly_kernel_method}
K_r(x,w)
=
(x^\top w)^r ,
\end{equation}
where $w\in\mathbb{R}^{d}$ denotes a kernel center. 
This atom represents degree-$r$ interactions in a compact kernel form, making it suitable for approximating the $r$-th order Volterra component.

Unlike classical kernel expansions, the centers in kVNN are not fixed training samples. 
They are learnable parameters optimized together with the network. 
Unlike Kervolution-like kernelized convolution~\cite{Wang2019CVPR}, the proposed representation is explicitly organized by Volterra order, so different interaction orders have separate centers and coefficients.

Therefore, kVNN can be viewed as a learnable, order-decoupled finite-atom Volterra-kernel representation. 
It preserves the ability of Volterra models to characterize higher-order interactions, avoids direct high-order tensor parameterization, and provides a compact operator that can be trained as part of a neural network layer.

\subsection{Geometric Interpretation}
\label{subsec:geometric_interpretation}

The geometric interpretation of kVNN progresses from a single cell to a layer and then to a multilayer network. 
A single kVNN cell provides a local higher-order approximation, a kVNN layer aggregates multiple local approximants, and a multilayer kVNN network composes such layer-level representations into a nonlinear model. Note that, due to space limitations, the proofs below are presented as proof sketches.

\subsubsection{Cell-Level Local Approximation}
\label{subsubsec:cell_level_approx}

The first level of interpretation concerns a single kVNN cell. Each kVNN cell provides a localized multi-kernel chart approximation, as formalized below.

\textbf{Lemma 1.}
Let $\Omega\subset\mathbb{R}^{d}$ be open, let $x_0\in\Omega$, and let $h:=g\circ\varphi\in C^{4}(\Omega)$. 
Then for every $\epsilon>0$, there exists $\delta>0$ such that, for every compact set $C\subset B(x_0,\delta)\cap\Omega$, there exists a finite kVNN expression
\begin{equation}
\label{eq:cell_level_kvnn_approx}
\begin{aligned}
F_{\epsilon}(x)
&=
b
+
\gamma_{1,1}(x^\top w_{1,1})
+
\sum_{j=1}^{r_2}
\gamma_{2,j}(x^\top w_{2,j})^2  \\
&\quad
+
\sum_{i=1}^{r_3}
\gamma_{3,i}(x^\top w_{3,i})^3 .
\end{aligned}
\end{equation}
satisfying
\begin{equation}
\label{eq:cell_level_kvnn_error}
\sup_{x\in C}
\left|
h(x)-F_{\epsilon}(x)
\right|
<\epsilon .
\end{equation}

\textit{Proof Sketch.}
Let $T_3$ be the third-order Taylor polynomial of $h$ at $x_0$. 
Since $h\in C^4(\Omega)$, Taylor's theorem gives
\[
h(x)=T_3(x)+R_4(x),
\]
where the remainder can be made uniformly smaller than $\epsilon$ on $C$ by choosing $\delta$ sufficiently small. 
Write $T_3(x)=c+\ell(x)+q(x)+r(x)$, where the terms are constant, linear, quadratic, and cubic, respectively. 
The constant and linear terms match $b+\gamma_{1,1}(x^\top w_{1,1})$. 
The quadratic and cubic terms can be represented as finite sums of powers of linear forms:
\[
q(x)=\sum_{j=1}^{r_2}\gamma_{2,j}(x^\top w_{2,j})^2,
\qquad
r(x)=\sum_{i=1}^{r_3}\gamma_{3,i}(x^\top w_{3,i})^3 .
\]
Thus, $T_3$ has the kVNN form in Eq.~\eqref{eq:cell_level_kvnn_approx}. 
Setting $F_{\epsilon}=T_3$ completes the proof. \hfill$\square$

As illustrated in Fig.~\ref{fig:kvnn_structure}, a kVNN operator is organized through order-specific learnable kernel branches. Each branch corresponds to one interaction order and is parameterized by its own centers and coefficients. This order-decoupled structure allows the number of paths to be selected separately for different orders: lower-order responses may use fewer paths, while higher-order responses can be assigned more paths for greater flexibility. Thus, the cell adapts its representation capacity across interaction orders while maintaining a compact finite-atom form\footnote{This Lemma can be readily extended to any order $p$.}.

\subsubsection{Layer-Level Multi-Chart Representation}
\label{subsubsec:layer_level_approx}

Each kVNN layer aggregates a finite collection of such local chart approximants into a multi-chart representation through linear mixing.

\textbf{Theorem 1.}
Let $\mathcal{M}\subset\mathbb{R}^{d}$ be compact and let $f\in C(\mathcal{M})$. 
Suppose that $\mathcal{M}$ admits a finite chart cover $\{V_\nu\}_{\nu=1}^{S}$, and that on each chart $V_\nu$, the local representative $f\circ\varphi_\nu^{-1}$ is sufficiently smooth so that Lemma~1 applies. 
Then for every $\epsilon>0$, there exist local kVNN approximants $\{F_\nu\}_{\nu=1}^{S}$ and a partition of unity $\{\rho_\nu\}_{\nu=1}^{S}$ subordinate to $\{V_\nu\}_{\nu=1}^{S}$ such that
\begin{equation}
\label{eq:layer_level_multichart}
\sup_{x\in\mathcal{M}}
\left|
f(x)
-
\sum_{\nu=1}^{S}
\rho_\nu(x)F_\nu(x)
\right|
<\epsilon .
\end{equation}

\textit{Proof Sketch.}
Since $\mathcal{M}$ is compact, the finite chart cover admits a partition of unity 
$\{\rho_\nu\}_{\nu=1}^{S}$ subordinate to $\{V_\nu\}_{\nu=1}^{S}$, with 
$\sum_{\nu=1}^{S}\rho_\nu(x)=1$. 
By Lemma~1, for each chart $V_\nu$, there exists a local kVNN approximant $F_\nu$ such that
$\sup_{x\in V_\nu\cap\mathcal{M}}|f(x)-F_\nu(x)|<\epsilon$. 
Define
\[
F(x)=\sum_{\nu=1}^{S}\rho_\nu(x)F_\nu(x).
\]
Then
\[
f(x)-F(x)
=
\sum_{\nu=1}^{S}\rho_\nu(x)\bigl(f(x)-F_\nu(x)\bigr),
\]
and therefore
\[
|f(x)-F(x)|
\leq
\sum_{\nu=1}^{S}\rho_\nu(x)|f(x)-F_\nu(x)|
<\epsilon .
\]
Taking the supremum over $x\in\mathcal{M}$ completes the proof. \hfill$\square$

At the implementation level, each kVNN cell computes an order-specific nonlinear filtering response, and multiple cells are arranged in parallel, similar to CNN filters. Therefore, kVNN changes the local filter operator while preserving the CNN-like layer organization. This allows kVNN to directly replace a convolutional layer and remain compatible with standard CNN design strategies such as depth-wise filtering and channel scaling. The structure of layer is shown in Fig.~\ref{fig:kvnn_structure}.

\subsubsection{Network-Level Compositional Approximation}
\label{subsubsec:network_level_approx}

The composition of layers constructs a global atlas whose transition maps are implicitly parameterized by inter-layer transformations.

\textbf{Theorem 2.}
Let $\mathcal{X}\subset\mathbb{R}^{d}$ be compact and let $f\in C(\mathcal{X})$. 
Consider a deep architecture formed by compositions of kVNN layers,
\[
\mathcal{N}_L
=
\mathcal{K}_L\circ\cdots\circ\mathcal{K}_1,
\]
where each layer contains finitely many kVNN cells together with linear mixing and sufficient localization. 
Then for every $\epsilon>0$, there exist a depth $L$ and network parameters such that
\[
\sup_{x\in\mathcal{X}}
\left|
f(x)-\mathcal{N}_L(x)
\right|
<\epsilon .
\]

\textit{Proof Sketch.}
Since $\mathcal{X}$ is compact, it admits a finite chart cover. 
By Theorem~1, the target function can be approximated by a finite multi-chart kVNN representation
\[
G(x)
=
\sum_{\nu=1}^{S}
\rho_\nu(x)F_\nu(x)
\]
with $\sup_{x\in\mathcal{X}}|f(x)-G(x)|<\epsilon/2$. 
A finite composition of kVNN layers with linear mixing and localization can realize this multi-chart representation up to error $\epsilon/2$, so that
\[
\sup_{x\in\mathcal{X}}
|G(x)-\mathcal{N}_L(x)|
<\epsilon/2 .
\]
Then the result follows from the triangle inequality. \hfill$\square$

\section{Experiments}
\label{sec:exp}

After establishing the formulation and structural interpretation, experiments evaluate whether the proposed operator improves the accuracy–efficiency trade-off in practical networks.

\subsection{Video Action Recognition}
\label{subsec:video_action}

Video action recognition is used as the main operator-level evaluation task, since it requires modeling both appearance and motion-related interactions. 
Experiments are conducted on UCF101~\cite{DBLP:journals/corr/abs-1212-0402}. 
To isolate the effect of the proposed operator, all compared methods are implemented under the same two-stream backbone and training protocol. 
RGB frames and optical-flow inputs are processed by two separate streams, and the resulting features are fused before classification. 
The two-stream model is used only as a controlled evaluation framework; the main comparison concerns the layer operator used inside the backbone.

Two model scales are considered, denoted as low-capacity (-L) and high-capacity (-H), which differ only in encoder depth. 
For kVNN, both second-order and third-order versions are evaluated. 
All models use the same training and testing settings, including a batch size of 16, 16-frame video clips, input resolution of $112\times112$, and Adam optimizer with learning rate $5\times10^{-4}$.

\begin{table}[t]
\centering
\caption{Comparison with Volterra- and kernel-based operators on UCF101 under two model scales. All models use the same backbone. Latency denotes the inference time of one forward pass, and Runtime is the time of one training step.}
{\small
\setlength{\tabcolsep}{3pt}
\begin{tabular}{lccccc}
\hline
Model & Params & GFLOPs & Latency & Runtime & Acc. \\
\hline
VNN-L~\cite{JMLR:v25:21-1082} & 12.4M & 28.48 & 0.1491s & 0.6800s & 86.16\% \\
Kervolution-L~\cite{Wang2019CVPR} & 7.2M & 15.31 & 0.1027s & 0.3887s & 84.77\% \\
kVNN-L (2nd) & 7.7M & 15.96 & 0.1043s & 0.3944s & 86.51\% \\
kVNN-L (3rd) & 12.2M & 28.14 & 0.1485s & 0.6788s & 90.02\% \\
\hline
VNN-H~\cite{JMLR:v25:21-1082} & 27.1M & 35.25 & 0.1629s & 0.7413s & 90.28\% \\
Kervolution-H~\cite{Wang2019CVPR} & 17.1M & 19.21 & 0.1310s & 0.6133s & 87.61\% \\
kVNN-H (2nd) & 17.8M & 19.86 & 0.1361s & 0.6293s & 91.17\% \\
kVNN-H (3rd) & 29.1M & 35.18 & 0.1604s & 0.7306s & 92.67\% \\
\hline
\end{tabular}
}
\label{tab:ucf101_comparison}
\end{table}

Table~\ref{tab:ucf101_comparison} shows that kVNN achieves a favorable accuracy--efficiency trade-off under both model scales. 
In the low-capacity setting, kVNN-L with second-order filters improves over Kervolution-L with only a small increase in parameters and GFLOPs, while also outperforming VNN-L with substantially lower complexity. 
Using third-order filters further improves accuracy to 90.02\%, exceeding VNN-L by 3.86 percentage points at comparable computational cost. 
In the high-capacity setting, kVNN-H with second-order filters outperforms both VNN-H and Kervolution-H, and kVNN-H with third-order filters achieves the best accuracy of 92.67\%. 
These results indicate that the proposed order-specific learnable kernel atoms improve representation capacity while maintaining controlled computational overhead.

\subsection{Image Denoising}
\label{subsec:image_denoising}

Image denoising is used to evaluate whether kVNN can serve as a plug-and-play replacement for convolutional filters in restoration networks. 
For each CNN baseline, convolutional blocks are replaced by kVNN blocks while keeping the overall backbone and training setting controlled. 
A single model is trained under random AWGN, where the noise level is sampled from $\mathcal{U}(0,50)$ for each training patch. 
Results are reported on Set12 at $\sigma=25$.

\begin{table}[t]
\centering
\caption{Random-$\sigma$ AWGN denoising on Set12 at $\sigma=25$. CNN baselines and their kVNN counterparts are compared under the same backbone settings. Latency denotes the inference time of one forward pass, and Runtime denotes the time of one training step.}
{\small
\setlength{\tabcolsep}{2.5pt}
\begin{tabular}{llccccc}
\hline
Method & Type & Params & GFLOPs & Latency & Runtime & PSNR \\
\hline
\multirow{2}{*}{DnCNN~\cite{10.1109/TIP.2017.2662206}}  
& CNN  & 0.55M & 2.285 & 0.0089s & 0.0340s & 30.31\\
& kVNN & 0.37M & 1.517 & 0.0071s & 0.0310s & 30.45\\
\hline
\multirow{2}{*}{FFDNet~\cite{10762fa0005c48318dd9aef22de37a1c}} 
& CNN  & 0.55M & 2.288 & 0.0091s & 0.0372s & 30.27\\
& kVNN & 0.37M & 1.517 & 0.0075s & 0.0331s & 30.41\\
\hline
\multirow{2}{*}{DCANet~\cite{10.1007/s00530-024-01469-8}} 
& CNN  & 1.4M & 5.106 & 0.1071s & 0.4195s & 30.47\\
& kVNN & 1.3M & 4.706 & 0.0972s & 0.3901s & 30.56\\
\hline
\end{tabular}
}
\label{tab:set12_randomsigma_sigma25}
\end{table}

\begin{figure}[t]
    \centering
    \includegraphics[width=0.9\linewidth]{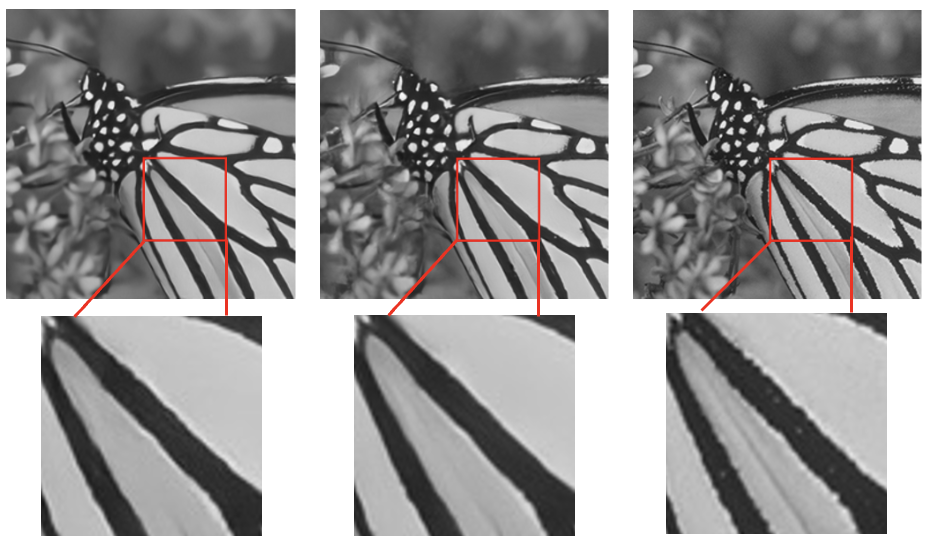}

    \vspace{1mm}

    \includegraphics[width=0.9\linewidth]{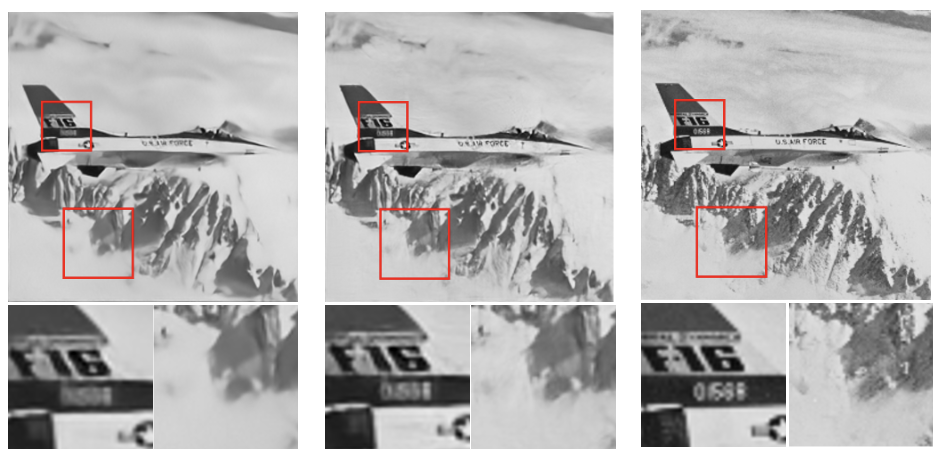}

    \caption{Visual comparison of Gaussian denoising results on Set12 at $\sigma=25$. 
    The examples compare the CNN-based model, the kVNN-based model, and the ground-truth image, with zoomed-in regions highlighting texture recovery and boundary preservation.}
    \label{fig:denoising_visual}
\end{figure}

Table~\ref{tab:set12_randomsigma_sigma25} shows that kVNN improves the accuracy--efficiency trade-off across different denoising backbones. 
For DnCNN and FFDNet, kVNN reduces the parameter count from 0.55M to 0.37M and lowers GFLOPs by about one third, while improving PSNR by 0.14 dB. 
For DCANet, kVNN also reduces GFLOPs, latency, and training runtime, while increasing PSNR from 30.47 dB to 30.56 dB. 
The visual comparisons in Fig.~\ref{fig:denoising_visual} further show that the kVNN-based model better preserves fine textures and object boundaries while suppressing residual noise. 
These results indicate that kVNN can serve as a compact replacement beyond the video setting, supporting its use as a general higher-order filtering operator.

\subsection{Image Classification}
\label{subsec:image_classification}

As an additional experiment, image classification is used to evaluate whether kVNN can also be integrated into CNN--Transformer hybrid recognition backbones. 
The Ti variant of Conformer~\cite{9709973} is adopted as the baseline. 
In the kVNN-based variant, the convolutional layers in the original backbone are replaced by kVNN layers, while the overall hybrid architecture and training setting are retained. 
To keep a favorable accuracy--efficiency trade-off, the number of layers in the kVNN-based variant is slightly reduced.

\begin{table}[t]
\centering
\caption{ImageNet-1K classification results using Conformer-Ti as the backbone. The kVNN variant replaces convolutional layers with kVNN layers while retaining the overall hybrid architecture.}
\label{tab:conformer_imagenet}
{\small
\setlength{\tabcolsep}{7pt}
\begin{tabular}{lcccc}
\hline
Model & Type & Params & MACs & Top-1 \\
\hline
Conformer-Ti~\cite{9709973} & CNN  & 23.5M & 5.2G & 81.3\% \\
Conformer-Ti & kVNN & 23.2M & 5.1G & 81.6\% \\
\hline
\end{tabular}
}
\end{table}

Table~\ref{tab:conformer_imagenet} shows that the kVNN variant slightly improves Top-1 accuracy from 81.3\% to 81.6\%, while also reducing parameter count and MACs. This result provides additional evidence that kVNN can be used beyond the video and denoising settings, including in CNN--Transformer hybrid architectures.

\section{Conclusion}
\label{sec:conclusion}

This paper presented kVNN, a learnable multi-kernel Volterra operator that combines order-wise Volterra filtering with polynomial-kernel atoms. The proposed formulation avoids explicit high-order tensor parameterization, removes the rigidity of fixed kernelized convolutional forms, and remains compatible with CNN-style layer construction. Its geometric interpretation further connects kVNN cells and layers with local and multi-chart approximation. Experiments on video action recognition, image denoising, and image classification demonstrate that kVNN provides an effective and compact higher-order filtering operator for signal processing tasks.

\bibliographystyle{IEEEbib}
\bibliography{strings,refs}

\end{document}